\documentclass{article}

\usepackage[preprint]{neurips_2025}

\usepackage[utf8]{inputenc} 
\usepackage[T1]{fontenc}    
\usepackage{hyperref}       
\usepackage{url}            
\usepackage{booktabs}       
\usepackage{amsfonts}       
\usepackage{nicefrac}       
\usepackage{microtype}      
\usepackage{xcolor}         
\usepackage{amsmath}
\usepackage{graphicx}
\usepackage{multirow}
\usepackage{array}

\title{Graph Fourier Transformer with Structure-Frequency Information}

\author{%
  Yonghui Zhai\\
  Dalian University of Technology\\
  Dalian, China\\
  \texttt{zhaiyonghui@mail.dlut.edu.cn} \\
  \And
  Yang Zhang\thanks{Corresponding author}\\
  Xidian University\\
  Xi'an, China\\
  \texttt{yangzhang1984@gmail.com}\\
  \AND
  Minghao Shang\\
  Xidian University\\
  Xi'an, China\\
  \texttt{shangmh1996@163.com}\\
  \And
  Lihua Pang\\
  Xi'an University of Science and Technology\\
  Xi'an, China\\
  \texttt{lhpang.xidian@gmail.com}\\
  \And
  Yaxin Ren\\
  Xidian University\\
  Xi'an, China\\
  \texttt{renyaxin.cn@gmail.com}\\
}
\begin{document}
\maketitle

\begin{abstract}
  Graph Transformers (GTs) have shown advantages in numerous graph structure tasks but their self-attention mechanism ignores the generalization bias of graphs, with existing methods mainly compensating for this bias from aspects like position encoding, attention bias and relative distance yet still having sub-optimal performance and being insufficient by only considering the structural perspective of generalization bias. To address this, this paper proposes Grafourierformer, which innovatively combines GT with inductive bias containing Frequency-Structure information by applying Graph Fourier Transform to the Attention Matrix: specifically, eigenvalues from the Graph Laplacian matrix are used to construct an Eigenvalue matrix mask (reflecting node positions and structural relationships with neighboring nodes to enable consideration of node range structural characteristics and focus on local graph details), and inverse Fourier transform is employed to extract node high-frequency and low-frequency features, calculate low-frequency and high-frequency energy, and construct a node frequency-energy matrix to filter the eigenvalue matrix mask, allowing attention heads to incorporate both graph structural information and node frequency information optimization, adaptively distinguish global trends from local details, and effectively suppress redundant information interference. Extensive experiments on various benchmarks show Grafourierformer consistently outperforms GNN and GT-based models in graph classification and node classification tasks, with ablation experiments further validating the effectiveness and necessity of the method. Codes are available at https://github.com/Arichibald/Grafourierformer.git
\end{abstract}
\section{Introduction}
Graph-structured data plays a pivotal role in modern machine learning, with applications spanning social networks, molecular biology, and recommender systems. Traditional Graph Neural Networks (GNNs) rely on message-passing mechanisms to aggregate local neighborhood information, but their limited receptive fields hinder their ability to capture complex global dependencies. Graph Transformers (GTs)~\citep{dwivedi2020generalization} address this limitation through self-attention mechanisms, enabling direct interactions between any pair of nodes and facilitating global information aggregation. However, the uniform attention mechanism in standard GTs fails to account for the inherent inductive biases in graph data—namely, structural correlations and frequency characteristics among nodes—resulting in suboptimal performance in capturing local structural details and multi-scale features.

Existing approaches attempt to mitigate this issue through two primary strategies: (1) positional encoding, where methods such as GT~\citep{dwivedi2020generalization} and SAN~\cite{kreuzer2021rethinking} employ Laplacian eigenvectors to encode structural information, and (2) attention bias, where models like Gradformer~\citep{liu2024gradformer} adjust attention scores based on the shortest path distances between nodes. However, as illustrated in Figure \ref{fig:Structure}, these methods exhibit only marginal differences in attention patterns compared to standard GTs, suggesting insufficient optimization of the attention mechanism. Crucially, these approaches focus solely on structural inductive bias while neglecting the frequency-domain properties of graphs—where low-frequency signals correspond to global trends and high-frequency signals reflect local details~\citep{shuman2013emerging}. This oversight renders existing models susceptible to redundant information interference, particularly in low-resource settings or complex graph structures, ultimately limiting their performance.

To address these limitations, we propose Graph Fourier Transformer (Grafourierformer), a novel architecture that integrates graph Fourier transforms into the GT framework, enabling joint structure-frequency modeling for multi-dimensional inductive bias injection. Our key innovations include: 1) Laplacian Eigenvalue Mask: Leveraging the eigenvalues of the graph Laplacian matrix to construct a structure-sensitive attention mask, explicitly modeling local connectivity patterns and global structural correlations. 2) Node Frequency Energy Filter: Decomposing node features into low- and high-frequency components via inverse Fourier transform, then dynamically adjusting attention weights based on their energy distribution to enhance multi-scale feature extraction.

This dual mechanism establishes a novel paradigm: "structural constraints guide attention range, while frequency optimization refines attention weights." Grafourierformer offers three key advantages: 1) Unified Modeling: The graph Fourier transform enables the first joint representation of structural and frequency-domain information, providing a more comprehensive inductive bias. 2) Adaptive Feature Separation: The frequency energy filter dynamically distinguishes between global trends and local details, effectively suppressing redundant information. 3) Hybrid Local-Global Learning: The combined structural-frequency mask preserves the local sensitivity while retaining the global modeling strength.

Extensive experiments on eight benchmark datasets demonstrate that Grafourierformer outperforms 15 baseline models across most tasks, with particularly significant gains in low-resource scenarios. For instance, on the NCI1 dataset with only 5\% training data, Grafourierformer achieves a 1.59\% accuracy improvement over the best competitor. These results validate the efficacy of joint structure-frequency modeling.

Contributions:

1.We propose the first GT framework integrating graph Fourier transforms, unifying structural and frequency-domain optimization.

2.We design Laplacian eigenvalue masks and frequency energy filters to inject multi-dimensional inductive biases into attention mechanisms.

3.We empirically demonstrate Grafourierformer’s superiority, particularly its robustness in low-resource settings.

The remainder of this paper is organized as follows: \autoref{sec:Methodology} details Grafourierformer’s methodology; \autoref{sec:Experiments} presents experimental results and analysis; \autoref{sec:Related Work} reviews related work; and \autoref{sec:Conclusion} concludes with future research directions.

\section{Methodology}
\label{sec:Methodology}
\subsection{Preliminaries}
Let an undirected graph be denoted as $G=(V,E)$, consisting of nodes $V$ and edges $E$. The graph $G$ can be represented by an adjacency matrix $A \in \{0,1\}^{n \times n}$ and a node feature matrix $X \in \mathbb{R}^{n \times d}$, where $n$ is the number of nodes, $d$ is the feature dimension, and $A[i,j] = 1$ if there exists an edge between nodes $v_i$ and $v_j$, otherwise $A[i,j] = 0$.

Traditional Graph Transformers (GTs) employ parameter matrices $W_Q^l$, $W_K^l$, and $W_V^l \in \mathbb{R}^{d_l \times d_{l+1}}$ to project node features into queries $Q^l$, keys $K^l$, and values $V^l$:

\begin{equation}
Q^l = x_{\text{node}}^l W_Q^l, \quad K^l = x_{\text{node}}^l W_K^l, \quad V^l = x_{\text{node}}^l W_V^l,
\end{equation}

where $l$ denotes the layer index of the Transformer and $d_l$ represents the hidden dimension at layer $l$.

The attention mechanism computes dot products between queries and keys to generate attention maps, subsequently producing output node embeddings $x_{\text{node}}^{l+1}$:

\begin{equation}
A^l = \frac{Q^l {K^l}^T}{\sqrt{d_l}}, \quad x_{\text{node}}^{l+1} = \text{softmax}(A^l) V^l,
\end{equation}

where $A^l \in \mathbb{R}^{n \times n \times d_{l+1}}$ represents the output attention map. This single-head self-attention module can be generalized to a multi-head self-attention (MHA) module through concatenation operations.

\subsection{Proposed Method: Grafourierformer}
This subsection describes the Grafourierformer method. First, we describe the architecture of the Grafourierformer and the process of graph Fourier transform, focusing our attention on two modules: the Laplace matrix eigenvalue mask and the node frequency energy filter. We explore in detail how the Grafourierformer method optimizes the self-attention matrix.

\paragraph{Architecture}
\begin{figure}
    \centering
    \includegraphics[width=1\linewidth]{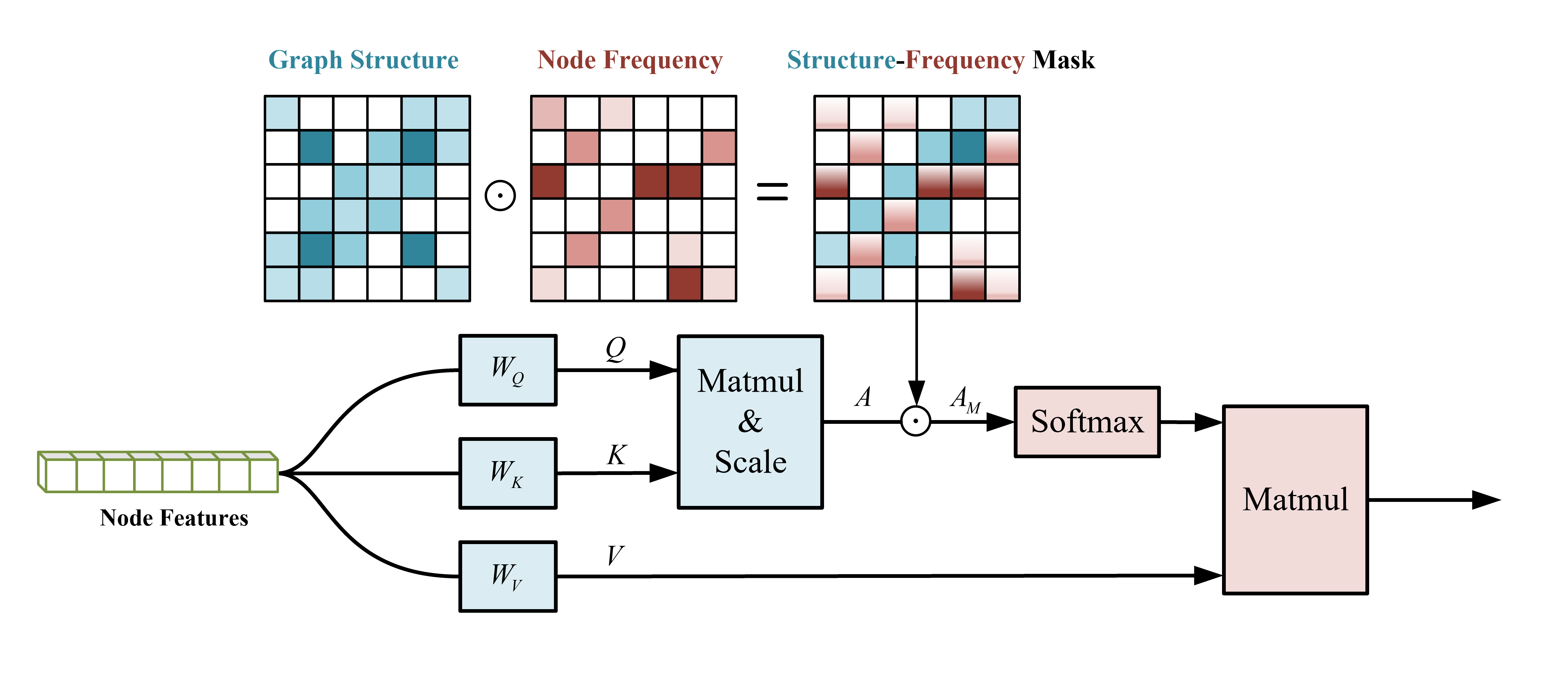}
    \caption{The structure of Grafourierformer is optimized by Structure-Frequency Mask on the attention matrix to achieve the fusion of graph structure information and node frequency information with attention.}
    \label{fig:Structure}
\end{figure}

As shown in Figure~\ref{fig:Structure}, Grafourierformer introduces Structure-Frequency Mask to optimize the self-attention matrix on the basis of GT, and injects the graph structure information and node frequency information into the self-attention matrix by using graph Fourier Transform, which introduces a more comprehensive inductive bias in self-attention learning.

\paragraph{Graph Fourier Transform}
Consider an undirected graph $G=(V,E)$ with adjacency matrix $A \in \{0,1\}^{n \times n}$, where $V$ represents the set of nodes with $|V|=N$ and $E$ denotes the set of edges. The normalized graph Laplacian matrix is defined as:

\begin{equation}
L = I_n - D^{-1/2} A D^{-1/2}
\end{equation}

where $D \in \mathbb{R}^{n \times n}$ is a diagonal degree matrix with elements $D_{i,i} = \sum_j A_{i,j}$, and $I_n$ denotes the identity matrix. As $L$ is a real symmetric matrix, it possesses a complete set of orthogonal eigenvectors ${u_l}_{l=1}^n \in \mathbb{R}^n$, each associated with a corresponding eigenvalue $\lambda_l \in [0,2]$~\citep{chung1997spectral}. Through eigendecomposition, we have: $L = U \Lambda U^T$, where $\Lambda = \text{diag}([\lambda_1, \lambda_2, \dots, \lambda_n])$.

According to graph signal processing theory~\citep{shuman2013emerging}, the eigenvectors of the normalized Laplacian matrix can be regarded as the basis for the graph Fourier transform. Given a node feature $v \in \mathbb{R}^n$, the graph Fourier transform is defined as $\hat{v} = U^T v$, and the inverse graph Fourier transform as $v = U \hat{v}$. Consequently, the convolution $*_G$ between node feature $v$ and convolution kernel $f$ can be expressed as:

\begin{equation}
f *_{G} v = U \left( (U^T f) \odot (U^T v) \right) = U g\theta U^T v
\end{equation}

where $\odot$ denotes the element-wise product of vectors, and $g_\theta$ is a diagonal matrix representing the convolution kernel in the spectral domain, which replaces $U^T f$.

\paragraph{Laplace Matrix Eigenvalue Mask}
\begin{figure}
    \centering
    \includegraphics[width=0.9\linewidth]{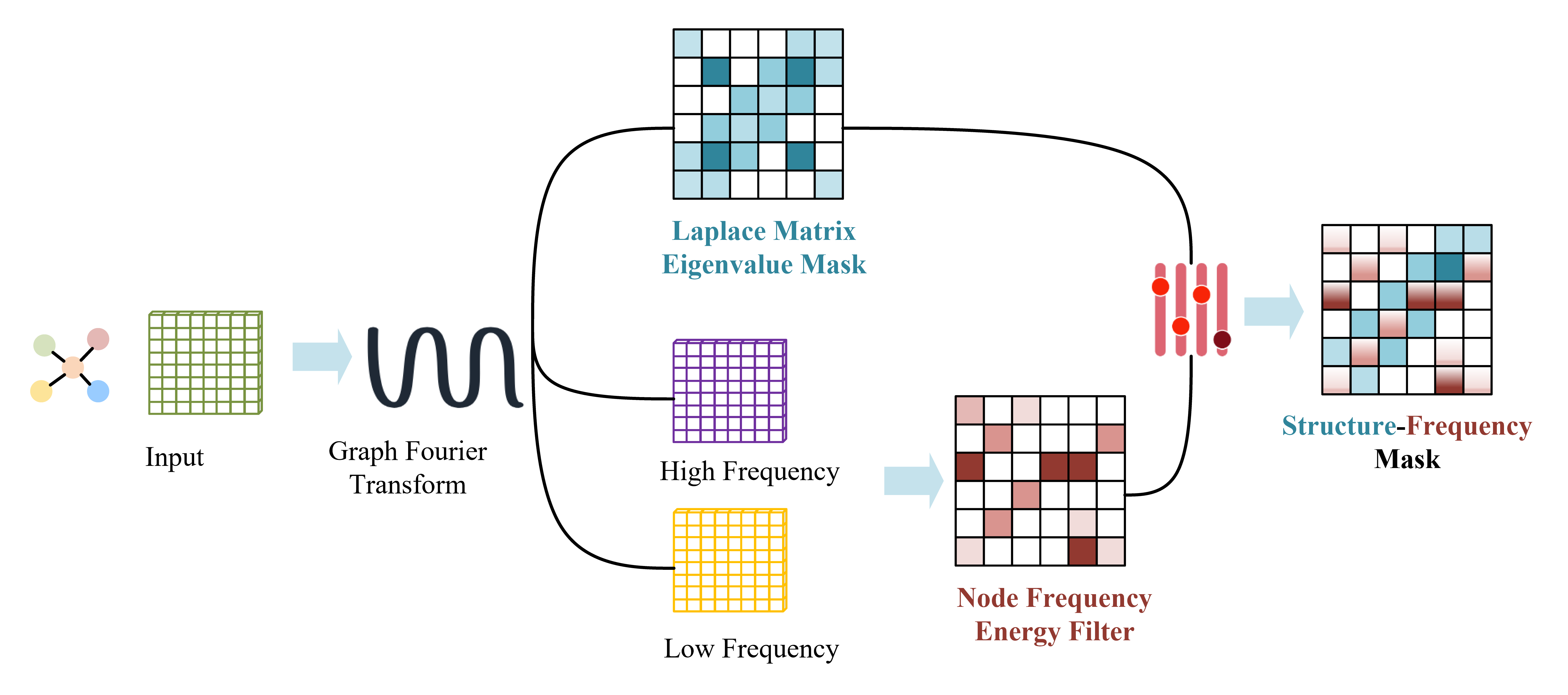}
    \caption{Figure Fourier filtering mechanism. The blue matrix denotes the Laplace Matrix Eigenvalue Mask, where the intensity of the blue color indicates the degree of the eigenvalue. The red matrix denotes the Filter Matrix, where the intensity of the red color indicates the size of the mask. After applying the mask (represented by the red cells with a gradient), the attentional values in the masked cells are significantly optimized. With this attention mask, the self-attention mechanism becomes more sensitive to the structure-frequency features of the graph.}
    \label{fig:Fourier}
\end{figure}
The eigenvalues of the graph Laplacian matrix reflect the connectivity properties of nodes in the graph. Smaller eigenvalues correspond to relatively sparsely connected components or isolated subgraphs in the graph. Conversely, larger eigenvalues are typically associated with densely connected regions where nodes have abundant edge connections, forming highly interconnected substructures. More details can be found in Appendix \ref{sec:proof}.

Therefore, we construct a Laplacian eigenvalue mask:

\begin{equation}
M_{ij} =
\begin{cases}
1, & \text{if } \lambda_i \leq 1 \\
0, & \text{if } \lambda_i > 1
\end{cases}
\end{equation}

where $S \in \mathbb{R}^{n \times n}$ is a structural information matrix with elements: $S_{i,j} = \lambda_i + \lambda_j$, $\lambda_i$ represents the eigenvalue corresponding to the Laplacian matrix. As the blue matrix in Figure \ref{fig:Fourier}, through this Laplacian eigenvalue mask construction, we effectively extract the structural information of nodes in the graph.

\paragraph{Node Frequency Energy Filter}
To fully leverage the complementary effects of low-frequency and high-frequency signals for node representation learning, we design a low-pass mask vector $\mathbf{m}_{\text{low}}$ and a high-pass mask vector $\mathbf{m}_{\text{high}}$ to achieve frequency signal separation from node features:

\begin{equation}
\mathbf{m}_{\text{low},i} = \begin{cases}
1, & \text{if } \lambda_i \leq 1 \\
0, & \text{if } \lambda_i > 1
\end{cases}, \quad
\mathbf{m}_{\text{high},i} = \begin{cases}
1, & \text{if } \lambda_i > 1 \\
0, & \text{if } \lambda_i \leq 1
\end{cases}
\end{equation}

Here, when the Laplacian eigenvalue $\lambda_i \leq 1$, $\mathbf{m}_{\text{low},i}=1$ (indicating low-frequency regions), otherwise 0; conversely, when $\lambda_i > 1$, $\mathbf{m}_{\text{high},i}=1$ (indicating high-frequency regions), otherwise 0. Through the inverse graph Fourier transform, the low-frequency node features can be expressed as:

\begin{equation}
\mathbf{v}_{\text{low}} = \mathbf{U}(\hat{\mathbf{v}} \odot \mathbf{m}_{\text{low}}), \quad \mathbf{v}_{\text{high}} = \mathbf{U}(\hat{\mathbf{v}} \odot \mathbf{m}_{\text{high}})
\end{equation}

where $\hat{\mathbf{v}}$ represents the frequency-domain features after Fourier transform, and $\mathbf{U}$ is the eigenvector matrix of the Laplacian matrix.

Furthermore, we construct low-frequency energy vector $\mathbf{e}_{\text{low}}$ and high-frequency energy vector $\mathbf{e}
_{\text{high}}$, where $\mathbf{e} \in \mathbb{R}^{1 \times n}$, by computing the energy distribution through squared summation of feature dimensions: 
\begin{equation}
    \mathbf{e}_{\text{low},i} = \sum_{k=1}^d \mathbf{v}_{\text{low},i,k}^2, \quad \mathbf{e}_{\text{high},i} = \sum_{k=1}^d \mathbf{v}_{\text{high},i,k}^2
\end{equation}
for $i = 1,2,\cdots,n$, where $d$ denotes the node feature dimension. The total energy $E$ is then calculated as:
\begin{equation}
    E = \sum_{i=1}^n \mathbf{e}_{\text{low},i} + \sum_{i=1}^n \mathbf{e}_{\text{high},i}
\end{equation}
 This total energy $E$, defined as the sum of all nodes' low-frequency and high-frequency energies, is used for normalization to construct the energy filter matrix $\mathbf{F}$: $\mathbf{F}_{i,j} = \frac{\mathbf{e}_{\text{low},i} + \mathbf{e}_{\text{high},j}}{E}$, This matrix dynamically adjusts the frequency sensitivity of attention weights by balancing the low-frequency energy of node $i$ with the high-frequency energy of node $j$. More node frequencies versus graphs are shown in Appendix \ref{sec:proof}.

Finally, the Fourier attention refinement matrix $\mathbf{M} = \text{ReLU}(\mathbf{S} \odot \mathbf{F})$, where $\mathbf{M} \in \mathbb{R}^{n \times n}$, is obtained through element-wise multiplication of the Laplacian eigenvalue mask $\mathbf{S}$ and the filter matrix $\mathbf{F}$. By injecting the Fourier attention refinement matrix $\mathbf{M}$ into the Transformer's attention matrix, As shown in Figure \ref{fig:Fourier}, the new attention matrix incorporates both structural and frequency information of nodes in the graph, achieving multi-perspective attention optimization:

\begin{equation}
\mathbf{A}_M^l = \frac{\mathbf{Q}^l {\mathbf{K}^l}^T \odot \mathbf{M}}{\sqrt{d_l}}, \quad \mathbf{x}_{\text{node}}^{l+1} = \text{softmax}(\mathbf{A}_M^l) \mathbf{V}^l
\end{equation}
\section{Experiments}
\label{sec:Experiments}
\subsection{Datasets}
We conduct comprehensive evaluations on eight widely-used real-world datasets to validate our model's effectiveness across diverse application scenarios. The benchmark includes: (1) two datasets from the GNN benchmark~\citep{dwivedi2023benchmarking} (PATTERN and CLUSTER), (2) five datasets from the TU repository~\citep{morris2020tudataset} (NCI1, PROTEINS, MUTAG, IMDB-BINARY, and COLLAB), and (3) the OGBG-MOLHIV dataset from OGB~\citep{hu2020open}. These datasets span multiple domains (including social networks, bioinformatics, synthetic graphs, and chemistry) and vary in scale (with OGBG-MOLHIV, PATTERN, and CLUSTER representing large-scale datasets). The evaluation covers both graph-level and node-level classification tasks. For all datasets, we strictly adhere to the standardized data splits and evaluation protocols established in~\citep{ying2021transformers}.

\subsection{Experimental Setup}
To rigorously validate the effectiveness of our proposed method, we conduct comprehensive comparisons between Grafourierformer and 15 state-of-the-art baselines, categorized as follows: (1) Four GCN-based models: GCN~\citep{kipf2016semi}, GAT~\citep{velivckovic2017graph}, GIN~\citep{xu2018powerful}, and GatedGCN~\citep{li2015gated}; (2) Eleven graph transformer variants: GraphTransformer~\citep{dwivedi2020generalization}, SAN~\citep{kreuzer2021rethinking}, Graphormer~\citep{ying2021transformers}, GraphTrans~\citep{wu2021representing}, SAT~\citep{chen2022structure}, EGT~\citep{hussain2022global}, GraphGPS~\citep{rampavsek2022recipe}, LGIGT~\citep{yin2023lgi}, KDLGT~\citep{wu2023kdlgt}, DeepGraph~\citep{zhao2023more}, and Gradformer~\citep{liu2024gradformer}. All baseline implementations strictly follow their official guidelines with recommended hyperparameter settings.

The evaluation framework employs the following protocol: (1) We assess model performance on both graph classification and regression tasks; (2) All implementations utilize the Adam optimizer within PyTorch; (3) For statistical reliability, we report mean performance metrics with standard deviations across 10 independent runs; (4) Model selection is based on optimal validation scores (Accuracy/MAE), with corresponding test set performance reported; (5) All experiments are conducted on servers equipped with dual NVIDIA RTX 4090 GPUs to ensure computational consistency.

\subsection{Overall Performance}
\begin{table}[htbp]
\centering
\caption{Experimental results for 8 common datasets (mean accuracy (Acc.) and AUROC and standard deviation for 10 different runs). \textbf{Bold}: best performance for each dataset. (*: DeepGraph uses a different evaluation metric on the pattern dataset than other models. Using this metric, our method achieves an accuracy of 90.67 ± 0.03. **: results are fine-tuned from a large pre-trained model.)}
\label{tab:Overall Performance}
\scalebox{0.65}{
\begin{tabular}{lcccccccc}
\toprule
Model & NCI & PROTE. & MUTAG & COLLAB & IMDB-B & PATTERN & CLUSTER & MOLHIV \\
& Acc. $\uparrow$ & Acc. $\uparrow$ & Acc. $\uparrow$ & Acc. $\uparrow$ & Acc. $\uparrow$ & Acc. $\uparrow$ & Acc. $\uparrow$ & AUROC $\uparrow$ \\
\midrule
\multicolumn{9}{l}{\textbf{GCN-based methods}} \\
GCN \citep{kipf2016semi} & 79.68±2.05 & 71.70±4.70 & 73.40±10.8 & 71.92±1.18 & 74.30±4.60 & 71.89±0.33 & 69.50±0.98 & 75.99±1.19 \\
GAT \citep{velivckovic2017graph} & 79.88±0.88 & 72.00±3.30 & 73.90±10.7 & 75.80±1.6 & 74.70±4.7 & 78.27±0.19 & 70.59±0.45 & - \\
GIN \citep{xu2018powerful} & 81.70±1.70 & 73.76±4.61 & 84.50±8.90 & 73.32±1.08 & 75.10±4.9 & 85.39±0.14 & 64.72±1.55 & 77.07±1.49 \\
GatedGCN \citep{li2015gated} & 81.17±0.79 & 74.65±1.13 & 85.00±2.67 & 80.70±0.75 & 73.20±1.32 & 85.57±0.09 & 73.84±0.33 & - \\
\midrule
\multicolumn{9}{l}{\textbf{Graph Transformer-based methods}} \\
GT \citep{dwivedi2020generalization} & 80.15±2.04 & 73.94±3.78 & 83.90±6.5 & 79.63±1.02 & 73.10±2.11 & 84.81±0.07 & 73.17±0.62 & - \\
SAN \citep{kreuzer2021rethinking} & 80.50±1.30 & 74.11±3.07 & 78.80±2.90 & 79.42±1.61 & 72.10±2.30 & 86.58±0.04 & 76.69±0.65 & 77.85±2.47 \\
GraphFormer \citep{ying2021transformers} & 81.44±0.57 & 75.29±3.10 & 80.52±5.79 & 81.80±2.24 & 73.40±2.80 & 86.65±0.03 & 74.66±0.24 & 74.55±1.06 \\
GraphTrans \citep{wu2021representing} & 82.60±1.20 & 75.18±3.36 & 87.22±7.05 & 79.81±0.84 & 74.50±2.89 & - & - & 76.33±1.11 \\
SAT \citep{chen2022structure} & 80.69±1.55 & 73.32±2.36 & 80.50±2.84 & 80.05±0.55 & 75.90±0.94 & 86.85±0.04 & 77.86±0.10 & - \\
EGT \citep{hussain2022global} & 81.91±3.42 & - & - & - & - & 86.82±0.02 & 79.23±0.18 & 80.51±0.30** \\
GraphGPS \citep{rampavsek2022recipe} & 84.21±2.25 & 75.77±2.19 & 85.00±3.16 & 81.40±1.26 & 77.40±0.63 & 86.69±0.06 & 78.02±0.18 & 78.80±0.49 \\
LGI-GT \citep{yin2023lgi} & 82.18±1.90 & - & - & - & - & 86.93±0.04 & 78.19±0.10 & - \\
KDLGT \citep{wu2023kdlgt} & - & - & - & - & - & - & 78.98±1.78 & - \\
DeepGraph \citep{zhao2023more} & - & - & - & - & - & 90.66±0.06* & 77.91±0.14 & - \\
Gradformer \citep{liu2024gradformer} & 86.01±1.47 & 77.50±1.86 & 88.00±2.45 & 82.01±1.06 & 77.10±0.54 & 86.89±0.07 & 78.55±0.16 & 79.15±0.89 \\
\midrule
Grafourierformer (Ours) & \textbf{87.13±0.43} & \textbf{79.55±3.04} & \textbf{89.84±3.42} & \textbf{82.40±0.86} & \textbf{79.25±1.68} & 86.69±0.03 & 77.26±0.34 & \textbf{80.28±1.77} \\
\bottomrule
\end{tabular}
}
\end{table}
We evaluate the effectiveness of our proposed model by comparing it with both GCN-based and GT-based models. For each model and dataset, we conduct 10 trials with random seeds and measure the average Acc/AUROC and standard deviation, as shown in Table 1. The experimental results in Table \ref{tab:Overall Performance} demonstrate that Grafourierformer achieves significant performance advantages on all eight benchmark datasets, attaining state-of-the-art (SOTA) performance on six datasets, which fully validates the effectiveness of our method. Specifically, on biomedical datasets (NCI1, PROTEINS, MUTAG) and social network datasets (IMDB-BINARY), Grafourierformer improves classification accuracy by 1.30\%, 2.65\%, 2.09\%, and 2.79\%, respectively, compared to the suboptimal baselines. These improvements stem from the model’s deep integration of structural information and frequency information into the self-attention mechanism through graph Fourier transforms, effectively enhancing its ability to model both local structural details and global dependencies. On large-scale datasets, although Grafourierformer’s performance is slightly lower than the optimal results, it remains competitive. Notably, the model shows more significant performance improvements on small datasets or in data-limited scenarios (see \autoref{subsec: Low-resource}), suggesting its strong adaptability to low- resource environments. Cross-domain experimental results show that Grafourierformer consistently achieves optimal or leading performance on tasks with different graph structural characteristics (e.g., node density, connectivity), demonstrating its broad applicability.

\subsection{Results on Low-resource Settings}
\label{subsec: Low-resource}
\begin{table}[htbp]
\centering
\caption{Results for low resource settings. 5\% , 10\%, 25\%, 100\% indicates that the model is trained using 5\%, 10\%, 25\%, and 100\% of the training set dataset, respectively. \textbf{Bold}: optimal performance for each data scale.}
\label{tab:Results on Low-resource Settings}
\begin{tabular}{lcccc}
\toprule
NCI1 & 5\% & 10\% & 25\% & 100\% \\
\midrule
GraphGPS & 69.54±0.96 & 74.70±0.44 & 76.55±1.19 & 84.21±2.25 \\
Gradformer & 71.20±0.49 & 76.38±0.85 & 77.98±1.22 & 86.01±1.47 \\
Ours & \textbf{72.33±0.75} & \textbf{76.94±0.39} & \textbf{79.34±1.47} & \textbf{87.13±0.43} \\
\bottomrule
\end{tabular}
\end{table}
To assess the model’s effectiveness in data-scarce scenarios, we conducted experiments on the NCI1 dataset using different proportions of training data (5\%, 10\%, 25\%, 100\%). The results in Table \ref{tab:Results on Low-resource Settings} show: 1) Grafourierformer significantly outperforms all baseline models across all data proportions. Particularly with only 5\% training data, it achieves a 1.59\% accuracy improvement over Gradformer, and shows gains of 0.73\% and 1.74\% with 10\% and 25\% data respectively. 2) This advantage stems from the model’s enhanced inductive bias through joint structure-frequency masking, which enables effective learning of essential graph characteristics from limited data, overcoming the limitations of conventional GTs that rely solely on structural information. The results further highlight Grafourierformer’s practical value in real-world scenarios with few-shot learning or high data annotation costs.

\subsection{Ablation Experiment}
\begin{table}[htbp]
\centering
\caption{GraphGPS, Graformer, and our model do not have the MPNN and PE modules on the NCI1 dataset, respectively, and the down arrow (\textcolor{red}{↓}) indicates a decrease in model performance relative to the baseline approach.}
\label{tab:ablation_study_1}
\begin{tabular}{lccc}
\toprule
\cmidrule(r){2-4} & GraphGPS & Gradformer & Ours \\
\midrule
- & 84.20 & 86.01 & 87.13 \\
w/o MPNN & 80.04 (\textcolor{red}{↓4.16}) & 80.80 (\textcolor{red}{↓5.21}) & 81.47 (\textcolor{red}{↓5.66}) \\
w/o PE & 83.67 (\textcolor{red}{↓0.53}) & 85.01 (\textcolor{red}{↓1.00}) & 85.04 (\textcolor{red}{↓2.09}) \\
\bottomrule
\end{tabular}
\end{table}

\begin{table}[htbp]g
\centering
\caption{The performance of our model without the MPNN, PE and Frequency Filter modules on the NCI1, PROTEINS, MUTAG and IMDB-BINARY datasets, respectively, with the down arrow (\textcolor{red}{↓}) indicating a decrease in model performance relative to the baseline approach.}
\label{tab:ablation_multi_dataset}
\begin{tabular}{lcccc}
\toprule
\cmidrule(r){2-5} & NCI1 & PROTEINS & MUTAG & IMDB-BINARY \\
\midrule
-  & 87.13 & 79.55 & 89.84 & 79.25 \\
w/o MPNN & 81.47 (\textcolor{red}{↓5.66}) & 73.83 (\textcolor{red}{↓5.72}) & 87.50 (\textcolor{red}{↓2.34}) & 75.90 (\textcolor{red}{↓3.35}) \\
w/o PE & 85.04 (\textcolor{red}{↓2.09}) & 75.69 (\textcolor{red}{↓3.86}) & 85.42 (\textcolor{red}{↓4.42}) & 77.00 (\textcolor{red}{↓2.25}) \\
w/o Frequencyfilter & 83.80 (\textcolor{red}{↓3.33}) & 76.55 (\textcolor{red}{↓3.00}) & 80.83 (\textcolor{red}{↓9.01}) & 78.95 (\textcolor{red}{↓0.30}) \\
\bottomrule
\end{tabular}
\end{table}
\begin{figure}
    \centering
    \includegraphics[width=0.6\linewidth]{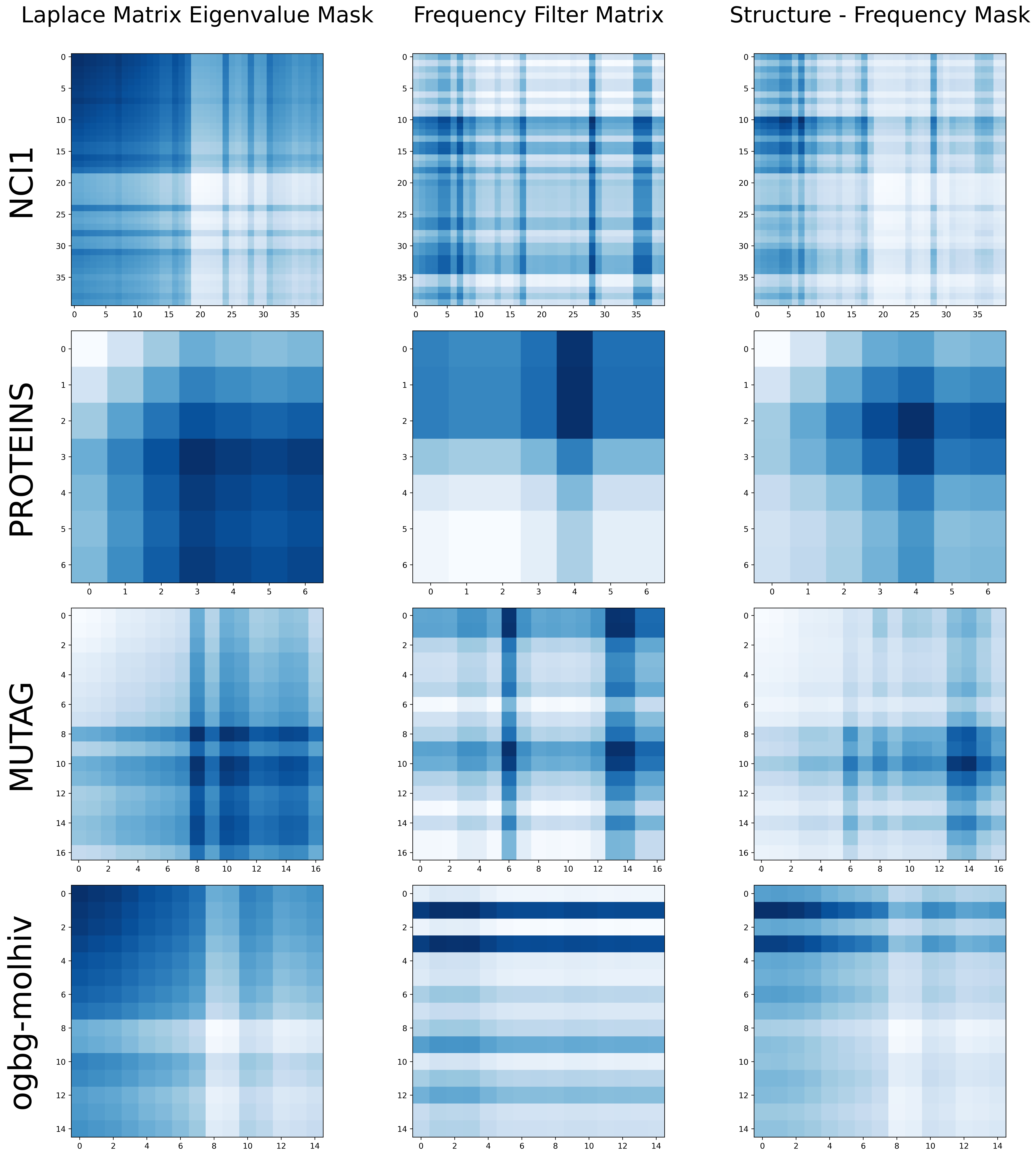}
    \caption{Laplace Matrix Eigenvalue Mask, Frequency Filter Matrix, and Structure-Frequency Mask visualizations on the NCI1, PROTEINS, MUTAG, and ogbg-molhiv datasets, respectively. (We choose randomly selected plots with the same ordinal number on each dataset as examples, with darker colors representing larger values)}
    \label{fig:matrix_plot}
\end{figure}
\begin{figure}
    \centering
    \includegraphics[width=0.8\linewidth]{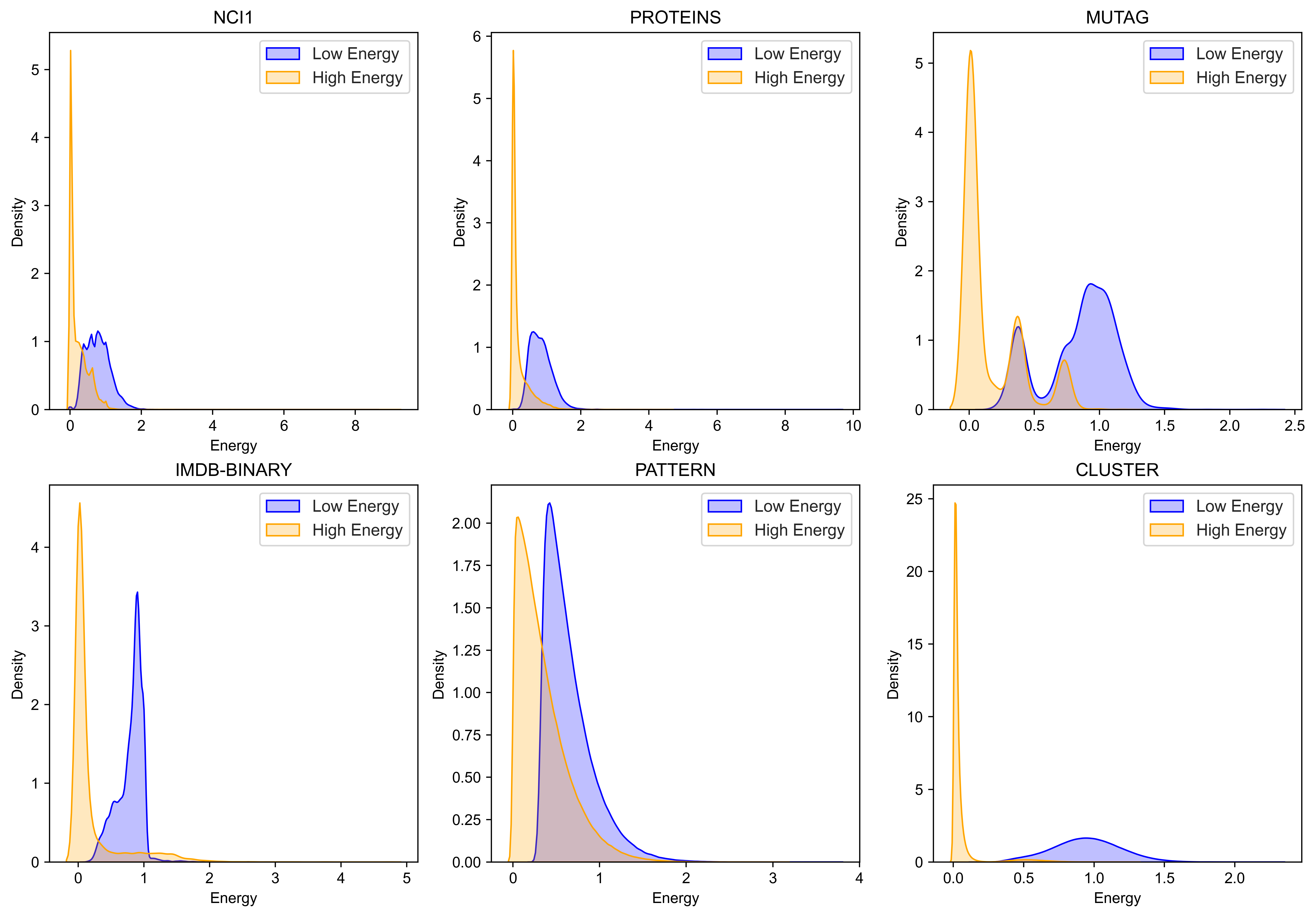}
    \caption{Kernel densities of high-frequency energies and low-frequency energies of nodes in the training sets of the NCI, PROTEINS, MUTAG, IMDB-BINARY, PATTERN, and CLUSTER datasets.}
    \label{fig:density}
\end{figure}
To validate key components, ablation studies on MPNN, PE, and Frequency Filter were conducted via horizontal (model) and vertical (dataset) comparisons.
Horizontal comparison (Table \ref{tab:ablation_study_1}): Removing MPNN/PE degraded all models' performance, but Grafourierformer retained the highest accuracy. For example, on NCI1, Grafourierformer's accuracy dropped by 5.66\% without MPNN, outperforming Gradformer (5.21\% drop) and GraphGPS (4.16\% drop), indicating graph Fourier transform enhances attention robustness.
Vertical comparison (Table \ref{fig:density}): Grafourierformer showed milder PE removal impacts on NCI1 (2.09\%) and IMDB-BINARY (2.25\%) vs. PROTEINS (3.86\%), suggesting frequency filters compensate for PE.
Removing Frequency Filter caused significant drops (e.g., 9.01\% on MUTAG), validating its critical role in attention optimization.
Results confirm that joint structure-frequency modeling drives Grafourierformer's performance via synergistic enhancement of self-attention for multi-dimensional graph feature capture.

\subsection{Visualization of Structure-Frequency Information}

To intuitively demonstrate the model’s capability in extracting graph features, we conducted qualitative and quantitative validation of structure-frequency information through matrix visualization and kernel density analysis. As shown in Figure \ref{fig:matrix_plot}, the Laplacian eigenvalue mask clearly reflects the structural correlations among nodes (e.g., high-value regions correspond to dense subgraphs). The frequency filter matrix dynamically modulates the structural mask based on low- and high-frequency energy distributions, enabling the joint structure-frequency mask to simultaneously capture local details (high-frequency) and global trends (low-frequency). As shown in Figure \ref{fig:density}, Low-frequency energy is significantly concentrated in high-value regions, indicating its dominant role in attention optimization and confirming the importance of low-frequency signals for global graph structure modeling; High-frequency energy primarily clusters in low-value regions, demonstrating that the frequency filter effectively suppresses redundant high-frequency noise while highlighting critical local features. In summary, the visualization results provide intuitive evidence that Grafourierformer successfully decouples and synergistically utilizes structure-frequency information, further validating the model’s design rationale.
\section{Related Work}
\label{sec:Related Work}
\paragraph{Graph Transformers} GTs have emerged as powerful alternatives to traditional GNNs by incorporating global self-attention mechanisms for graph representation learning. Initial work by \cite{dwivedi2020generalization} established the foundation by extending Transformers to arbitrary graph structures using Laplacian eigenvectors as positional encodings. Subsequent developments include molecular graph-specific architectures like \cite{chen2021learning}'s CoMPT, multi-scale feature extraction in \cite{zhu2021posegtac}'s PoseGTAC, and structural bias incorporation in \cite{chen2022structure}'s SAT. Scalability improvements have been achieved through methods like \cite{choromanski2022block}'s block-Toeplitz masks and node-level optimizations \citep{liu2024exploring,sun2023retentive}, with applications spanning multilingual processing \cite{ding2020self}, knowledge graphs \cite{cai2022entity}, and spatiotemporal prediction \cite{wu2023difformer}.

\paragraph{Prior Knowledge in Graph Transformer} Prior knowledge integration in GTs follows these main directions: enhanced positional encodings (e.g., SAN's learnable eigenfunctions \cite{kreuzer2021rethinking}), attention modifications (e.g., Graphormer's structural biases \cite{ying2021transformers} and EGT's edge features \cite{hussain2022global}), and architectural hybrids (e.g., GraphGPS's parallel GNN-transformer design \cite{rampavsek2022recipe}). Key innovations include Graphormer's triple structural encoding, SAN's spectral-aware PEs, EGT's edge-enhanced attention, NodeFormer's efficient structure learning \cite{wu2022nodeformer}, and GraphGPS's unified framework, collectively demonstrating that effective GT design requires spectral PEs, explicit edge modeling, and balanced local-global architectures.
\section{Conclusion}
\label{sec:Conclusion}
datasets demonstrate that Grafourierformer significantly outperforms 15 existing GNN and GT baseline models, validating its effectiveness and advancement. Notably, the core advantage of Grafourierformer lies in its multi-dimensional inductive bias injection mechanism: the model realizes dynamic optimization of the attention mechanism through structure-frequency joint masking, which not only improves the accuracy of the model in small-scale datasets or low-resource scenarios compared to contrast models but also achieves performance breakthroughs in complex graph structure tasks. Compared with traditional GT, Grafourierformer maintains optimal or highly competitive results in cross-domain generalization capabilities. It is worth emphasizing that the model's efficient learning ability in low-resource scenarios provides a new paradigm for practical applications with high data annotation costs and scarce samples.
Although Grafourierformer has demonstrated significant advantages, there are still directions for further exploration: 1) Fine-grained modeling of frequency information: Investigate the deeper correlation between high-low frequency energy distribution and graph structure features to further enhance the expressive power of the attention mechanism; 2) Efficiency optimization for large-scale graphs: Explore the potential of using only graph Fourier transform to optimize attention, reducing dependence on traditional message-passing modules to make it more suitable for graph data with large node scales.


\bibitem[Dwivedi and Bresson(2020)]{dwivedi2020generalization}
Vijay~Prakash Dwivedi and Xavier Bresson.
\newblock A generalization of transformer networks to graphs.
\newblock \emph{arXiv preprint arXiv:2012.09699}, 2020.

\bibitem[Kreuzer et~al.(2021)Kreuzer, Beaini, Hamilton, L{\'e}tourneau, and Tossou]{kreuzer2021rethinking}
Devin Kreuzer, Dominique Beaini, Will Hamilton, Vincent L{\'e}tourneau, and Prudencio Tossou.
\newblock Rethinking graph transformers with spectral attention.
\newblock \emph{Advances in Neural Information Processing Systems}, 34:\penalty0 21618--21629, 2021.

\bibitem[Liu et~al.(2024{\natexlab{a}})Liu, Yao, Zhan, Ma, Pan, and Hu]{liu2024gradformer}
Chuang Liu, Zelin Yao, Yibing Zhan, Xueqi Ma, Shirui Pan, and Wenbin Hu.
\newblock Gradformer: Graph transformer with exponential decay.
\newblock \emph{arXiv preprint arXiv:2404.15729}, 2024{\natexlab{a}}.

\bibitem[Shuman et~al.(2013)Shuman, Narang, Frossard, Ortega, and Vandergheynst]{shuman2013emerging}
David~I Shuman, Sunil~K Narang, Pascal Frossard, Antonio Ortega, and Pierre Vandergheynst.
\newblock The emerging field of signal processing on graphs: Extending high-dimensional data analysis to networks and other irregular domains.
\newblock \emph{IEEE signal processing magazine}, 30\penalty0 (3):\penalty0 83--98, 2013.

\bibitem[Chung(1997)]{chung1997spectral}
Fan~RK Chung.
\newblock \emph{Spectral graph theory}, volume~92.
\newblock American Mathematical Soc., 1997.

\bibitem[Dwivedi et~al.(2023)Dwivedi, Joshi, Luu, Laurent, Bengio, and Bresson]{dwivedi2023benchmarking}
Vijay~Prakash Dwivedi, Chaitanya~K Joshi, Anh~Tuan Luu, Thomas Laurent, Yoshua Bengio, and Xavier Bresson.
\newblock Benchmarking graph neural networks.
\newblock \emph{Journal of Machine Learning Research}, 24\penalty0 (43):\penalty0 1--48, 2023.

\bibitem[Morris et~al.(2020)Morris, Kriege, Bause, Kersting, Mutzel, and Neumann]{morris2020tudataset}
Christopher Morris, Nils~M Kriege, Franka Bause, Kristian Kersting, Petra Mutzel, and Marion Neumann.
\newblock Tudataset: A collection of benchmark datasets for learning with graphs.
\newblock \emph{arXiv preprint arXiv:2007.08663}, 2020.

\bibitem[Hu et~al.(2020)Hu, Fey, Zitnik, Dong, Ren, Liu, Catasta, and Leskovec]{hu2020open}
Weihua Hu, Matthias Fey, Marinka Zitnik, Yuxiao Dong, Hongyu Ren, Bowen Liu, Michele Catasta, and Jure Leskovec.
\newblock Open graph benchmark: Datasets for machine learning on graphs.
\newblock \emph{Advances in neural information processing systems}, 33:\penalty0 22118--22133, 2020.

\bibitem[Ying et~al.(2021)Ying, Cai, Luo, Zheng, Ke, He, Shen, and Liu]{ying2021transformers}
Chengxuan Ying, Tianle Cai, Shengjie Luo, Shuxin Zheng, Guolin Ke, Di~He, Yanming Shen, and Tie-Yan Liu.
\newblock Do transformers really perform badly for graph representation?
\newblock \emph{Advances in neural information processing systems}, 34:\penalty0 28877--28888, 2021.

\bibitem[Kipf and Welling(2016)]{kipf2016semi}
Thomas~N Kipf and Max Welling.
\newblock Semi-supervised classification with graph convolutional networks.
\newblock \emph{ICLR}, 2016.

\bibitem[Veli{\v{c}}kovi{\'c} et~al.(2017)Veli{\v{c}}kovi{\'c}, Cucurull, Casanova, Romero, Lio, and Bengio]{velivckovic2017graph}
Petar Veli{\v{c}}kovi{\'c}, Guillem Cucurull, Arantxa Casanova, Adriana Romero, Pietro Lio, and Yoshua Bengio.
\newblock Graph attention networks.
\newblock \emph{arXiv preprint arXiv:1710.10903}, 2017.

\bibitem[Xu et~al.(2018)Xu, Hu, Leskovec, and Jegelka]{xu2018powerful}
Keyulu Xu, Weihua Hu, Jure Leskovec, and Stefanie Jegelka.
\newblock How powerful are graph neural networks?
\newblock \emph{arXiv preprint arXiv:1810.00826}, 2018.

\bibitem[Li et~al.(2015)Li, Tarlow, Brockschmidt, and Zemel]{li2015gated}
Yujia Li, Daniel Tarlow, Marc Brockschmidt, and Richard Zemel.
\newblock Gated graph sequence neural networks.
\newblock \emph{arXiv preprint arXiv:1511.05493}, 2015.

\bibitem[Wu et~al.(2021)Wu, Jain, Wright, Mirhoseini, Gonzalez, and Stoica]{wu2021representing}
Zhanghao Wu, Paras Jain, Matthew Wright, Azalia Mirhoseini, Joseph~E Gonzalez, and Ion Stoica.
\newblock Representing long-range context for graph neural networks with global attention.
\newblock \emph{Advances in neural information processing systems}, 34:\penalty0 13266--13279, 2021.

\bibitem[Chen et~al.(2022)Chen, O’Bray, and Borgwardt]{chen2022structure}
Dexiong Chen, Leslie O’Bray, and Karsten Borgwardt.
\newblock Structure-aware transformer for graph representation learning.
\newblock In \emph{International conference on machine learning}, pages 3469--3489. PMLR, 2022.

\bibitem[Hussain et~al.(2022)Hussain, Zaki, and Subramanian]{hussain2022global}
Md~Shamim Hussain, Mohammed~J Zaki, and Dharmashankar Subramanian.
\newblock Global self-attention as a replacement for graph convolution.
\newblock In \emph{Proceedings of the 28th ACM SIGKDD Conference on Knowledge Discovery and Data Mining}, pages 655--665, 2022.

\bibitem[Ramp{\'a}{\v{s}}ek et~al.(2022)Ramp{\'a}{\v{s}}ek, Galkin, Dwivedi, Luu, Wolf, and Beaini]{rampavsek2022recipe}
Ladislav Ramp{\'a}{\v{s}}ek, Michael Galkin, Vijay~Prakash Dwivedi, Anh~Tuan Luu, Guy Wolf, and Dominique Beaini.
\newblock Recipe for a general, powerful, scalable graph transformer.
\newblock \emph{Advances in Neural Information Processing Systems}, 35:\penalty0 14501--14515, 2022.

\bibitem[Yin and Zhong(2023)]{yin2023lgi}
Shuo Yin and Guoqiang Zhong.
\newblock Lgi-gt: Graph transformers with local and global operators interleaving.
\newblock In \emph{IJCAI}, pages 4504--4512, 2023.

\bibitem[Wu et~al.(2023{\natexlab{a}})Wu, Xu, Zhu, Song, Lin, Wang, and Liu]{wu2023kdlgt}
Yi~Wu, Yanyang Xu, Wenhao Zhu, Guojie Song, Zhouchen Lin, Liang Wang, and Shaoguo Liu.
\newblock Kdlgt: A linear graph transformer framework via kernel decomposition approach.
\newblock In \emph{IJCAI}, pages 2370--2378, 2023{\natexlab{a}}.

\bibitem[Zhao et~al.(2023)Zhao, Ma, Zhang, Deng, and Wei]{zhao2023more}
Haiteng Zhao, Shuming Ma, Dongdong Zhang, Zhi-Hong Deng, and Furu Wei.
\newblock Are more layers beneficial to graph transformers?
\newblock \emph{arXiv preprint arXiv:2303.00579}, 2023.

\bibitem[Chen et~al.(2021)Chen, Zheng, Song, Rao, and Yang]{chen2021learning}
Jianwen Chen, Shuangjia Zheng, Ying Song, Jiahua Rao, and Yuedong Yang.
\newblock Learning attributed graph representations with communicative message passing transformer.
\newblock \emph{arXiv preprint arXiv:2107.08773}, 2021.

\bibitem[Zhu et~al.(2021)Zhu, Xu, Shen, Ji, Gao, and Shen]{zhu2021posegtac}
Yiran Zhu, Xing Xu, Fumin Shen, Yanli Ji, Lianli Gao, and Heng~Tao Shen.
\newblock Posegtac: Graph transformer encoder-decoder with atrous convolution for 3d human pose estimation.
\newblock In \emph{IJCAI}, pages 1359--1365, 2021.

\bibitem[Choromanski et~al.(2022)Choromanski, Lin, Chen, Zhang, Sehanobish, Likhosherstov, Parker-Holder, Sarlos, Weller, and Weingarten]{choromanski2022block}
Krzysztof Choromanski, Han Lin, Haoxian Chen, Tianyi Zhang, Arijit Sehanobish, Valerii Likhosherstov, Jack Parker-Holder, Tamas Sarlos, Adrian Weller, and Thomas Weingarten.
\newblock From block-toeplitz matrices to differential equations on graphs: towards a general theory for scalable masked transformers.
\newblock In \emph{International Conference on Machine Learning}, pages 3962--3983. PMLR, 2022.

\bibitem[Liu et~al.(2024{\natexlab{b}})Liu, Zhan, Ma, Ding, Tao, Wu, Hu, and Du]{liu2024exploring}
Chuang Liu, Yibing Zhan, Xueqi Ma, Liang Ding, Dapeng Tao, Jia Wu, Wenbin Hu, and Bo~Du.
\newblock Exploring sparsity in graph transformers.
\newblock \emph{Neural Networks}, 174:\penalty0 106265, 2024{\natexlab{b}}.

\bibitem[Sun et~al.(2023)Sun, Dong, Huang, Ma, Xia, Xue, Wang, and Wei]{sun2023retentive}
Yutao Sun, Li~Dong, Shaohan Huang, Shuming Ma, Yuqing Xia, Jilong Xue, Jianyong Wang, and Furu Wei.
\newblock Retentive network: A successor to transformer for large language models.
\newblock \emph{arXiv preprint arXiv:2307.08621}, 2023.

\bibitem[Ding et~al.(2020)Ding, Wang, and Tao]{ding2020self}
Liang Ding, Longyue Wang, and Dacheng Tao.
\newblock Self-attention with cross-lingual position representation.
\newblock \emph{arXiv preprint arXiv:2004.13310}, 2020.

\bibitem[Cai et~al.(2022)Cai, Ma, Zhan, and Jiang]{cai2022entity}
Weishan Cai, Wenjun Ma, Jieyu Zhan, and Yuncheng Jiang.
\newblock Entity alignment with reliable path reasoning and relation-aware heterogeneous graph transformer.
\newblock \emph{arXiv preprint arXiv:2205.08806}, 2022.

\bibitem[Wu et~al.(2023{\natexlab{b}})Wu, Yang, Zhao, He, Wipf, and Yan]{wu2023difformer}
Qitian Wu, Chenxiao Yang, Wentao Zhao, Yixuan He, David Wipf, and Junchi Yan.
\newblock Difformer: Scalable (graph) transformers induced by energy constrained diffusion.
\newblock \emph{arXiv preprint arXiv:2301.09474}, 2023{\natexlab{b}}.

\bibitem[Wu et~al.(2022)Wu, Zhao, Li, Wipf, and Yan]{wu2022nodeformer}
Qitian Wu, Wentao Zhao, Zenan Li, David Wipf, and Junchi Yan.
\newblock Nodeformer: A scalable graph structure learning transformer for node classification.
\newblock In \emph{Advances in Neural Information Processing Systems}, 2022.

\begin{thebibliography}{29}
\providecommand{\natexlab}[1]{#1}
\providecommand{\url}[1]{\texttt{#1}}
\expandafter\ifx\csname urlstyle\endcsname\relax
\providecommand{\doi}[1]{doi: #1}\else
\providecommand{\doi}{doi: \begingroup \urlstyle{rm}\Url}\fi

\bibitem[Dwivedi and Bresson(2020)]{dwivedi2020generalization}
Vijay~Prakash Dwivedi and Xavier Bresson.
\newblock A generalization of transformer networks to graphs.
\newblock \emph{arXiv preprint arXiv:2012.09699}, 2020.

\bibitem[Kreuzer et~al.(2021)Kreuzer, Beaini, Hamilton, L{\'e}tourneau, and Tossou]{kreuzer2021rethinking}
Devin Kreuzer, Dominique Beaini, Will Hamilton, Vincent L{\'e}tourneau, and Prudencio Tossou.
\newblock Rethinking graph transformers with spectral attention.
\newblock \emph{Advances in Neural Information Processing Systems}, 34:\penalty0 21618--21629, 2021.

\bibitem[Liu et~al.(2024{\natexlab{a}})Liu, Yao, Zhan, Ma, Pan, and Hu]{liu2024gradformer}
Chuang Liu, Zelin Yao, Yibing Zhan, Xueqi Ma, Shirui Pan, and Wenbin Hu.
\newblock Gradformer: Graph transformer with exponential decay.
\newblock \emph{arXiv preprint arXiv:2404.15729}, 2024{\natexlab{a}}.

\bibitem[Shuman et~al.(2013)Shuman, Narang, Frossard, Ortega, and Vandergheynst]{shuman2013emerging}
David~I Shuman, Sunil~K Narang, Pascal Frossard, Antonio Ortega, and Pierre Vandergheynst.
\newblock The emerging field of signal processing on graphs: Extending high-dimensional data analysis to networks and other irregular domains.
\newblock \emph{IEEE signal processing magazine}, 30\penalty0 (3):\penalty0 83--98, 2013.

\bibitem[Chung(1997)]{chung1997spectral}
Fan~RK Chung.
\newblock \emph{Spectral graph theory}, volume~92.
\newblock American Mathematical Soc., 1997.

\bibitem[Dwivedi et~al.(2023)Dwivedi, Joshi, Luu, Laurent, Bengio, and Bresson]{dwivedi2023benchmarking}
Vijay~Prakash Dwivedi, Chaitanya~K Joshi, Anh~Tuan Luu, Thomas Laurent, Yoshua Bengio, and Xavier Bresson.
\newblock Benchmarking graph neural networks.
\newblock \emph{Journal of Machine Learning Research}, 24\penalty0 (43):\penalty0 1--48, 2023.

\bibitem[Morris et~al.(2020)Morris, Kriege, Bause, Kersting, Mutzel, and Neumann]{morris2020tudataset}
Christopher Morris, Nils~M Kriege, Franka Bause, Kristian Kersting, Petra Mutzel, and Marion Neumann.
\newblock Tudataset: A collection of benchmark datasets for learning with graphs.
\newblock \emph{arXiv preprint arXiv:2007.08663}, 2020.

\bibitem[Hu et~al.(2020)Hu, Fey, Zitnik, Dong, Ren, Liu, Catasta, and Leskovec]{hu2020open}
Weihua Hu, Matthias Fey, Marinka Zitnik, Yuxiao Dong, Hongyu Ren, Bowen Liu, Michele Catasta, and Jure Leskovec.
\newblock Open graph benchmark: Datasets for machine learning on graphs.
\newblock \emph{Advances in neural information processing systems}, 33:\penalty0 22118--22133, 2020.

\bibitem[Ying et~al.(2021)Ying, Cai, Luo, Zheng, Ke, He, Shen, and Liu]{ying2021transformers}
Chengxuan Ying, Tianle Cai, Shengjie Luo, Shuxin Zheng, Guolin Ke, Di~He, Yanming Shen, and Tie-Yan Liu.
\newblock Do transformers really perform badly for graph representation?
\newblock \emph{Advances in neural information processing systems}, 34:\penalty0 28877--28888, 2021.

\bibitem[Kipf and Welling(2016)]{kipf2016semi}
Thomas~N Kipf and Max Welling.
\newblock Semi-supervised classification with graph convolutional networks.
\newblock \emph{ICLR}, 2016.

\bibitem[Veli{\v{c}}kovi{\'c} et~al.(2017)Veli{\v{c}}kovi{\'c}, Cucurull, Casanova, Romero, Lio, and Bengio]{velivckovic2017graph}
Petar Veli{\v{c}}kovi{\'c}, Guillem Cucurull, Arantxa Casanova, Adriana Romero, Pietro Lio, and Yoshua Bengio.
\newblock Graph attention networks.
\newblock \emph{arXiv preprint arXiv:1710.10903}, 2017.

\bibitem[Xu et~al.(2018)Xu, Hu, Leskovec, and Jegelka]{xu2018powerful}
Keyulu Xu, Weihua Hu, Jure Leskovec, and Stefanie Jegelka.
\newblock How powerful are graph neural networks?
\newblock \emph{arXiv preprint arXiv:1810.00826}, 2018.

\bibitem[Li et~al.(2015)Li, Tarlow, Brockschmidt, and Zemel]{li2015gated}
Yujia Li, Daniel Tarlow, Marc Brockschmidt, and Richard Zemel.
\newblock Gated graph sequence neural networks.
\newblock \emph{arXiv preprint arXiv:1511.05493}, 2015.

\bibitem[Wu et~al.(2021)Wu, Jain, Wright, Mirhoseini, Gonzalez, and Stoica]{wu2021representing}
Zhanghao Wu, Paras Jain, Matthew Wright, Azalia Mirhoseini, Joseph~E Gonzalez, and Ion Stoica.
\newblock Representing long-range context for graph neural networks with global attention.
\newblock \emph{Advances in neural information processing systems}, 34:\penalty0 13266--13279, 2021.

\bibitem[Chen et~al.(2022)Chen, O’Bray, and Borgwardt]{chen2022structure}
Dexiong Chen, Leslie O’Bray, and Karsten Borgwardt.
\newblock Structure-aware transformer for graph representation learning.
\newblock In \emph{International conference on machine learning}, pages 3469--3489. PMLR, 2022.

\bibitem[Hussain et~al.(2022)Hussain, Zaki, and Subramanian]{hussain2022global}
Md~Shamim Hussain, Mohammed~J Zaki, and Dharmashankar Subramanian.
\newblock Global self-attention as a replacement for graph convolution.
\newblock In \emph{Proceedings of the 28th ACM SIGKDD Conference on Knowledge Discovery and Data Mining}, pages 655--665, 2022.

\bibitem[Ramp{\'a}{\v{s}}ek et~al.(2022)Ramp{\'a}{\v{s}}ek, Galkin, Dwivedi, Luu, Wolf, and Beaini]{rampavsek2022recipe}
Ladislav Ramp{\'a}{\v{s}}ek, Michael Galkin, Vijay~Prakash Dwivedi, Anh~Tuan Luu, Guy Wolf, and Dominique Beaini.
\newblock Recipe for a general, powerful, scalable graph transformer.
\newblock \emph{Advances in Neural Information Processing Systems}, 35:\penalty0 14501--14515, 2022.

\bibitem[Yin and Zhong(2023)]{yin2023lgi}
Shuo Yin and Guoqiang Zhong.
\newblock Lgi-gt: Graph transformers with local and global operators interleaving.
\newblock In \emph{IJCAI}, pages 4504--4512, 2023.

\bibitem[Wu et~al.(2023{\natexlab{a}})Wu, Xu, Zhu, Song, Lin, Wang, and Liu]{wu2023kdlgt}
Yi~Wu, Yanyang Xu, Wenhao Zhu, Guojie Song, Zhouchen Lin, Liang Wang, and Shaoguo Liu.
\newblock Kdlgt: A linear graph transformer framework via kernel decomposition approach.
\newblock In \emph{IJCAI}, pages 2370--2378, 2023{\natexlab{a}}.

\bibitem[Zhao et~al.(2023)Zhao, Ma, Zhang, Deng, and Wei]{zhao2023more}
Haiteng Zhao, Shuming Ma, Dongdong Zhang, Zhi-Hong Deng, and Furu Wei.
\newblock Are more layers beneficial to graph transformers?
\newblock \emph{arXiv preprint arXiv:2303.00579}, 2023.

\bibitem[Chen et~al.(2021)Chen, Zheng, Song, Rao, and Yang]{chen2021learning}
Jianwen Chen, Shuangjia Zheng, Ying Song, Jiahua Rao, and Yuedong Yang.
\newblock Learning attributed graph representations with communicative message passing transformer.
\newblock \emph{arXiv preprint arXiv:2107.08773}, 2021.

\bibitem[Zhu et~al.(2021)Zhu, Xu, Shen, Ji, Gao, and Shen]{zhu2021posegtac}
Yiran Zhu, Xing Xu, Fumin Shen, Yanli Ji, Lianli Gao, and Heng~Tao Shen.
\newblock Posegtac: Graph transformer encoder-decoder with atrous convolution for 3d human pose estimation.
\newblock In \emph{IJCAI}, pages 1359--1365, 2021.

\bibitem[Choromanski et~al.(2022)Choromanski, Lin, Chen, Zhang, Sehanobish, Likhosherstov, Parker-Holder, Sarlos, Weller, and Weingarten]{choromanski2022block}
Krzysztof Choromanski, Han Lin, Haoxian Chen, Tianyi Zhang, Arijit Sehanobish, Valerii Likhosherstov, Jack Parker-Holder, Tamas Sarlos, Adrian Weller, and Thomas Weingarten.
\newblock From block-toeplitz matrices to differential equations on graphs: towards a general theory for scalable masked transformers.
\newblock In \emph{International Conference on Machine Learning}, pages 3962--3983. PMLR, 2022.

\bibitem[Liu et~al.(2024{\natexlab{b}})Liu, Zhan, Ma, Ding, Tao, Wu, Hu, and Du]{liu2024exploring}
Chuang Liu, Yibing Zhan, Xueqi Ma, Liang Ding, Dapeng Tao, Jia Wu, Wenbin Hu, and Bo~Du.
\newblock Exploring sparsity in graph transformers.
\newblock \emph{Neural Networks}, 174:\penalty0 106265, 2024{\natexlab{b}}.

\bibitem[Sun et~al.(2023)Sun, Dong, Huang, Ma, Xia, Xue, Wang, and Wei]{sun2023retentive}
Yutao Sun, Li~Dong, Shaohan Huang, Shuming Ma, Yuqing Xia, Jilong Xue, Jianyong Wang, and Furu Wei.
\newblock Retentive network: A successor to transformer for large language models.
\newblock \emph{arXiv preprint arXiv:2307.08621}, 2023.

\bibitem[Ding et~al.(2020)Ding, Wang, and Tao]{ding2020self}
Liang Ding, Longyue Wang, and Dacheng Tao.
\newblock Self-attention with cross-lingual position representation.
\newblock \emph{arXiv preprint arXiv:2004.13310}, 2020.

\bibitem[Cai et~al.(2022)Cai, Ma, Zhan, and Jiang]{cai2022entity}
Weishan Cai, Wenjun Ma, Jieyu Zhan, and Yuncheng Jiang.
\newblock Entity alignment with reliable path reasoning and relation-aware heterogeneous graph transformer.
\newblock \emph{arXiv preprint arXiv:2205.08806}, 2022.

\bibitem[Wu et~al.(2023{\natexlab{b}})Wu, Yang, Zhao, He, Wipf, and Yan]{wu2023difformer}
Qitian Wu, Chenxiao Yang, Wentao Zhao, Yixuan He, David Wipf, and Junchi Yan.
\newblock Difformer: Scalable (graph) transformers induced by energy constrained diffusion.
\newblock \emph{arXiv preprint arXiv:2301.09474}, 2023{\natexlab{b}}.

\bibitem[Wu et~al.(2022)Wu, Zhao, Li, Wipf, and Yan]{wu2022nodeformer}
Qitian Wu, Wentao Zhao, Zenan Li, David Wipf, and Junchi Yan.
\newblock Nodeformer: A scalable graph structure learning transformer for node classification.
\newblock In \emph{Advances in Neural Information Processing Systems}, 2022.

\end{thebibliography}
\end{document}